\documentclass[nohyperref]{article}

\usepackage{microtype}
\usepackage{graphicx}
\usepackage{booktabs} 

\usepackage{hyperref}

\usepackage[accepted]{icml2023}

\usepackage{amsmath}
\usepackage{amssymb}
\usepackage{mathtools}
\usepackage{amsthm}
\usepackage{float}

\usepackage[capitalize,noabbrev]{cleveref}

\theoremstyle{plain}

\theoremstyle{definition}

\theoremstyle{remark}

\usepackage[textsize=tiny]{todonotes}

\usepackage{siunitx}
\usepackage{svg}
\usepackage{comment}
\usepackage{subfig}

\DeclareSIUnit\angstrom{\text {Å}}

\icmltitlerunning{CryoChains}

\begin{document}

\twocolumn[
\icmltitle{CryoChains: Heterogeneous Reconstruction of Molecular Assembly of Semi-flexible Chains from Cryo-EM Images
}


\icmlsetsymbol{equal}{*}

\begin{icmlauthorlist}
\icmlauthor{Bongjin Koo}{ucsb}
\icmlauthor{Julien Martel}{stanford}
\icmlauthor{Ariana Peck}{lcls-slac}
\icmlauthor{Axel Levy}{stanford,lcls-slac}
\icmlauthor{Frédéric Poitevin}{lcls-slac}
\icmlauthor{Nina Miolane}{ucsb}
\end{icmlauthorlist}

\icmlaffiliation{ucsb}{Dept. of Electrical \& Computer Engineering, UC Santa Barbara, CA, USA}
\icmlaffiliation{lcls-slac}{LCLS, SLAC, Menlo Park, CA, USA}
\icmlaffiliation{stanford}{Stanford University, CA, USA}

\icmlcorrespondingauthor{Bongjin Koo}{bongjinkoo@ucsb.edu}

\icmlkeywords{Machine Learning, ICML}

\vskip 0.3in
]



\printAffiliationsAndNotice{}  

\begin{abstract}
Cryogenic electron microscopy (cryo-EM) has transformed structural biology by allowing to reconstruct 3D biomolecular structures up to near-atomic resolution. However, the 3D reconstruction process remains challenging, as the 3D structures may exhibit substantial shape variations, while the 2D image acquisition suffers from a low signal-to-noise ratio, requiring to acquire very large datasets that are time-consuming to process.
Current reconstruction methods are precise but computationally expensive, or faster but lack a physically-plausible model of large molecular shape variations. To fill this gap,
we propose CryoChains that encodes large deformations of biomolecules via rigid body transformation of their chains, while representing their finer shape variations with the normal mode analysis framework of biophysics.
Our synthetic data experiments on the human GABA\textsubscript{B} and heat shock protein show that CryoChains gives a biophysically-grounded quantification of the heterogeneous conformations of biomolecules, while reconstructing their 3D molecular structures at an improved resolution compared to the current fastest, interpretable deep learning method.
\end{abstract} 

\section{Introduction}
\label{sec:intro}

\begin{figure*}
  \centering
  \includegraphics[width=1\linewidth]{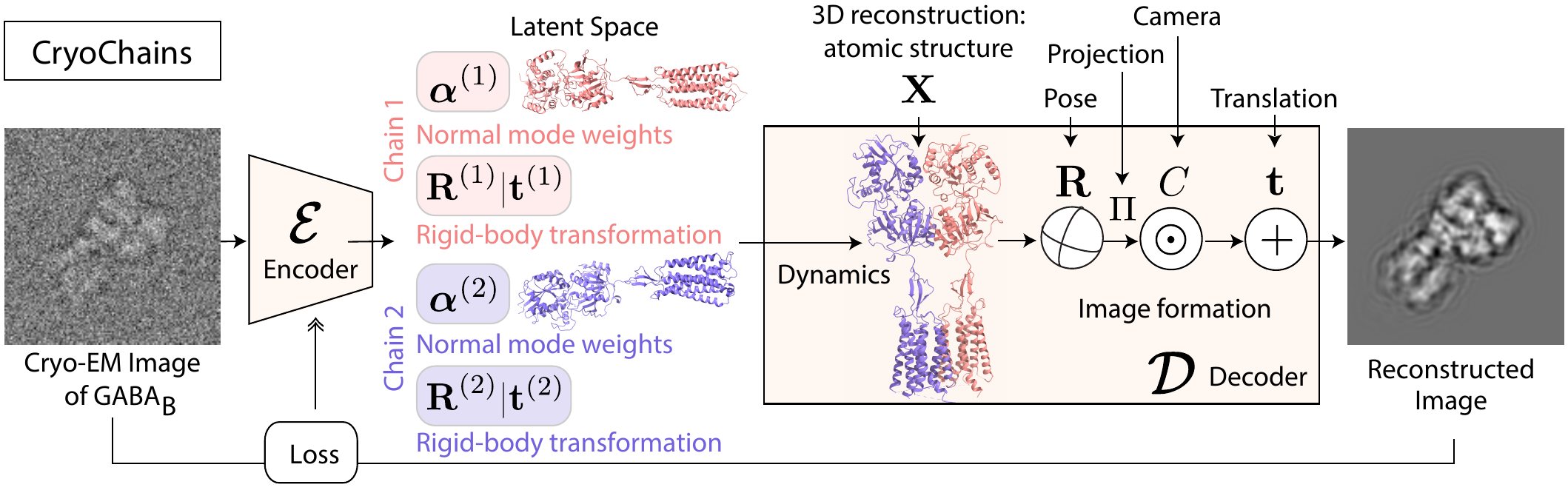}
  \caption{CryoChains autoencoder illustrated on the GABA\textsubscript{B} protein with two chains in red and blue. The encoder $\mathcal{E}$ embeds the input image into a latent space with two sets of 
  variables: (\textit{i}) $\boldsymbol{\alpha}^{(c)}$, the normal mode weights per chain, which deform each chain; and (\textit{ii}) $\textbf{R}^{(c)}|\textbf{t}^{(c)}$, the rigid body transformation that poses each (deformed) chain within the global atomic structure $\textbf{X}$ of the protein. The dynamics model of the decoder $\mathcal{D}$ applies the NMA deformations and rigid body transformations to each chain of a biomolecule, which is fed into the image formation model composed of a global rotation $\textbf{R}$ (``pose''), projection to 2D space ($\Pi$), camera parameters ($C$) and a global 2D translation $\textbf{t}$, to reconstruct the input image. CryoChains is trained using a reconstruction loss, which is the pixel-wise mean squared error between the input and reconstructed images.}
  \label{fig:cryochains_overview}
\end{figure*}

Cryogenic electron microscopy (cryo-EM) allows to
reconstruct 3D biomolecular structures at near-atomic resolution. For this, estimation techniques process large datasets of 2D images obtained from many copies of the same biomolecule in different shapes, called \textit{conformations}. 
The reconstruction task is challenging due to the intrinsic variability in a biomolecule conformations (\textit{conformation heterogeneity}), 
the high number of unknown, nuisance variables
(e.g., the biomolecule's position/orientation in 3D), and extremely low signal-to-noise ratios (SNRs).
Thus, reconstruction algorithms require large amounts of data,
posing major computational challenges~\cite{Kimanius2016AcceleratedRELION-2}.

\paragraph{Related works}
Recent efforts have turned to deep learning (DL) for cryo-EM reconstruction, reaching significant speed-ups 
through gradient methods and GPU acceleration~\cite{donnatDeepGenerativeModeling2022}. However, these approaches have been limited in their parameterization of the biomolecular volumes' variability. On the one hand, \textit{homogeneous reconstruction} methods disregard the conformation heterogeneity entirely, as they estimate 
the average 3D biomolecular volume from the images
~\cite{nashedCryoPoseNetEndtoEndSimultaneous2021a, MiolanePoitevin2020CVPR, gupta2021cryogan}, which limits the accuracy of the 3D reconstruction as an average biomolecule appears ``blurry''.
On the other hand, \textit{heterogeneous reconstruction} methods explicitly model
structural variability~\cite{ZhongCryoDRGNReconstructionHeterogeneous2021, zhong2021exploring, gupta2020multi, Ullrich2020, Chen2021, Rosenbaum2021, punjani3DFlexibleRefinement2022} but rarely leverage domain knowledge
to encode biophysically-plausible molecular deformations, which limits the interpretation of the conformations.



In biophysics, normal mode analysis (NMA) is a dimensionality reduction method,
describing the vibrational modes of a biomolecular structure, 
i.e., the principal directions of its conformational variations~\cite{levitt1985protein}. Thus, NMA has naturally emerged as the first biophysically-grounded approach to efficiently parameterize the biomolecular volumes and their variability for cryo-EM reconstruction. HEMNMA~\cite{Jin2014} uses normal modes to deform a reference volume $\textbf{V}^{(0)}$ that is fit to
cryo-EM images by an
iterative 3D-to-2D alignment.
DeepHEMNMA~\cite{hamitoucheDeepHEMNMAResNetbasedHybrid2022} extends HEMNMA
by adopting a residual network that predicts NMA weights (the contribution of each normal mode) 
and biomolecular poses given by HEMNMA in a supervised
learning setting. By contrast,
\cite{nashed_heterogeneous_2022}
performs unsupervised heterogeneous reconstruction, inferring NMA weights as the latent variables of an autoencoder. 
However, NMA can only represent small conformation changes around a reference conformation. 
This poses a fundamental limit as biomolecular variability exhibits larger scale deformations. For example, the rotations of a biomolecule's chains w.r.t. one another are typically larger deformations that are not captured by NMA.


\paragraph{Contributions} We propose a new modular, biophysically-grounded method to 
parameterize the molecular volumes' variability for accurate, interpretable heterogeneous reconstruction in cryo-EM. 
We decompose molecules into their \textit{chains} and represent
large scale molecular deformations by the rotation 
and translation of chains w.r.t. another. 
Then, we represent the smaller scale deformations per chain by NMA.
Building on \cite{nashed_heterogeneous_2022}, 
we propose a new unsupervised DL architecture \emph{CryoChains}, which extracts the rotation and translation, with 
the normal mode weights, of each chain in the latent space of an autoencoder (Fig.~\ref{fig:cryochains_overview}).
CryoChains achieves higher reconstruction accuracy on synthetic cryo-EM images with realistic noise levels,
compared to
the existing method using NMA only. We also show that CryoChains provides a biophysical interpretation of the conformation heterogeneity via the exploration of its latent space.

\begin{figure*}
\centering
  \includegraphics[width=1\linewidth]{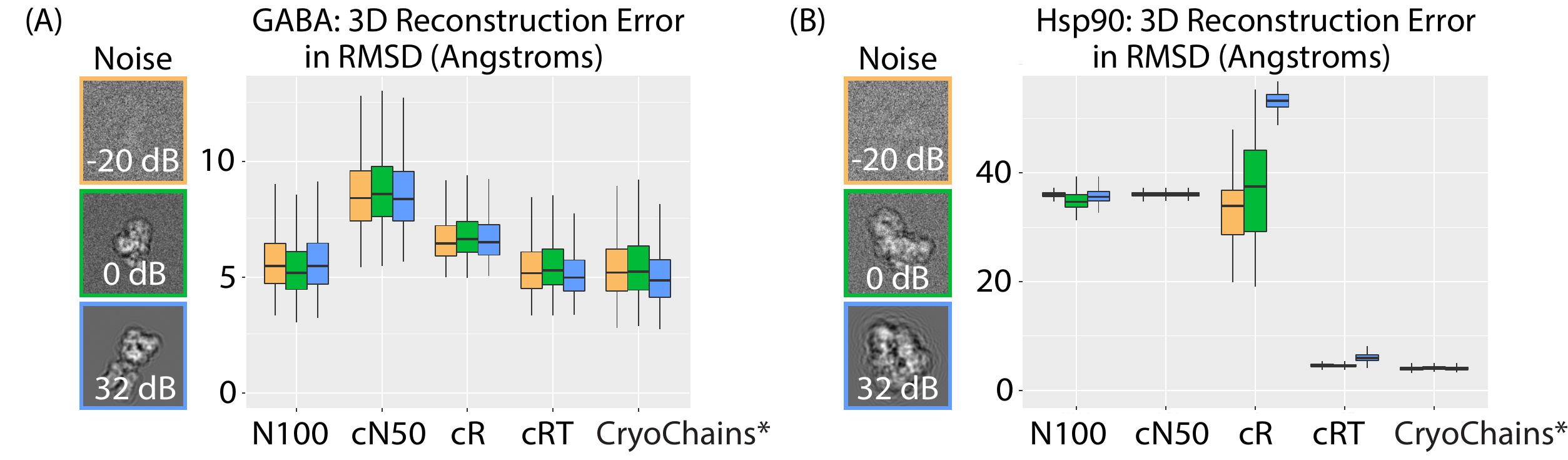}

\caption{Comparison of 3D reconstruction accuracy between CryoChains, its ablation models (\texttt{cN50, cR, cRT}), and the baseline~\texttt{N100}~\cite{nashed_heterogeneous_2022}. Reconstruction error is measured as the root-mean-square deviation (RMSD) in Angstroms, between the ground truth (GT) and predicted atom coordinates on GABA\textsubscript{B} and Hsp90, for 3 noise levels $\{32, 0, -20\}$ in dB shown in 3 colors. The models differ in their representation of conformation heterogeneity in the latent space: \texttt{N100} uses whole protein NMA with 100 normal modes, \texttt{cN50}: per-chain NMA with 50 normal modes, \texttt{cR}: per-chain rotation, \texttt{cRT}: per-chain rigid body transformation, CryoChains: ours, i.e., per-chain rigid body transformation and per-chain NMA with 50 normal modes. 
}
\label{fig:gaba-synthetic-experiments}
\end{figure*}

\section{CryoChains Methods}
\label{sec:method}

CryoChains adopts an autoencoder architecture shown in Fig.~\ref{fig:cryochains_overview}. The encoder $\mathcal{E}$ is a convolutional network that embeds a given image in a latent space composed of several conformation latent variables. 
The decoder $\mathcal{D}$ implements the generative model of cryo-EM images composed of the dynamics and image formation. This section and Appendix~\ref{sec:ae} detail $\mathcal{E}$ and $\mathcal{D}$.
The pixel-wise mean squared error (MSE) between the input and the reconstructed images is minimized during training.
Only the encoder is trained, i.e., the decoder is pre-specified and fixed.

\subsection{Encoder and Latent Space}
\label{subsec:nma}


The encoder $\mathcal{E}$ first extracts the small, elastic conformational variations of the biomolecule via a latent variable $\boldsymbol{\alpha}^{(c)}$ for each chain $c$\textemdash see Fig.~\ref{fig:cryochains_overview}. The variable $\boldsymbol{\alpha}^{(c)} \in \mathbb{R}^{K^{(c)}}$ represents the chain's deformation along its normal modes:
\begin{equation}
\label{eq:conformation_deform}
\textbf{X}^{(c)}(\boldsymbol{\alpha}) = \textbf{X}^{(c, 0)}+ \sum_{k = 1}^{K^{(c)}} \alpha^{(c)}_k \textbf{U}_k^{(c)}.
\end{equation}
Here, $\textbf{X}^{(c)}(\boldsymbol{\alpha}) \in \mathbb{R}^{3n_c}$ is the atomic structure of chain $c$ as the 3D coordinates of its $n_c$ constitutive atoms. NMA defines this atomic structure as the deformation of a reference structure $\textbf{X}^{(c, 0)} \in \mathbb{R}^{3n_c}$ through $K^{(c)}$ normal modes $\textbf{U}_1^{(c)}, \dots, \textbf{U}_{K^{(c)}}^{(c)}$ which are computed beforehand. 

Importantly, the weights $\boldsymbol{\alpha}^{(c)}$ along the normal modes have a natural biophysical interpretation. As 
normal modes are the eigenvectors of the Hessian matrix of the atoms' potential energy, obtained from a second order
Taylor approximation around a conformation at equilibrium, the weights encode in which direction and how far each atom moves
w.r.t. its position in the reference conformation\textemdash see~\cite{levitt1985protein,goldstein_classical_2008} and Appendix~\ref{sec:nma} for details.
The normal modes are sorted by the eigenvalues of the Hessian matrix in an increasing order, 
where the square-root of each eigenvalue is the frequency that deforms 
the reference conformation $\textbf{X}^{(0)}$ along each normal mode $\textbf{U}_{k}$. 
Thus, the first few normal modes describe low-frequency (large) conformation changes,
in which we are mainly interested. The number of normal modes $K^{(c)}$ per chain is chosen to be
the minimal number of normal modes
that could account for the majority of 
the deformations. 
We highlight that we use NMA on each chain independently,
as opposed to NMA on the whole biomolecule as proposed in~\cite{nashed_heterogeneous_2022}.


$\mathcal{E}$ also extracts the larger, rigid body transformations that arise 
when chains of a biomolecule rotate and translate w.r.t. each other.
We parameterize the orientation of each chain as a rotation
around its center-of-mass (CoM) to which we add a translation.
$\mathcal{E}$ extracts two 3D vectors that are orthonormalized into 
$\textbf{w}_{1}$ and $\textbf{w}_{2}$, from which 
the rotation matrix $\textbf{R}^{(c)}$ is built per chain.
Translation $\textbf{t}^{(c)}$ is a 3D vector per chain. We highlight that these per-chain rotation $\textbf{R}^{(c)}$ and translation $\textbf{t}^{(c)}$ differ from the global rotation $\textbf{R}$ and translation $\textbf{t}$ of the whole biomolecule within the decoder.


\subsection{Decoder}
\label{subsec:image_formation_model}


The dynamics component of the decoder $\mathcal{D}$ takes the conformation latent variables $\{\boldsymbol{\alpha}^{(c)}, \textbf{R}^{(c)}, \textbf{t}^{(c)}\}_{c=1}^{N_C}$ to output the 3D atomic structure $\textbf{X}$ of the biomolecule in the input 2D image\textemdash see Fig.~\ref{fig:cryochains_overview}. The structure $\textbf{X}^{(c)}$ of each chain $c$ is given by Eq.~(\ref{eq:conformation_deform}) and $\textbf{X}$ is given by
    $\textbf{X} = [\textbf{R}^{(c)}(\textbf{X}^{(c)}(\boldsymbol{\alpha}^{(c)})) + \textbf{t}^{(c)}]_{c=1}^{N_C}$,
where $[~]$ means concatenation.

The image formation component of the decoder $\mathcal{D}$ generates cryo-EM images using the atomic model $\textbf{X}$ given by the dynamics.
It is a feed-forward differentiable simulator without learnable components that successively creates a electrostatic potential $\textbf{V}$ from $\textbf{X}$, orients it in 3D space using a global rotation $\textbf{R}$, then acquires a 2D image through a camera model with parameter $C$, and finally adds a 2D translation $\textbf{t}$.
See Appendix~\ref{sec:image_formation_model} for details.

\begin{figure*}
  \centering
  \includegraphics[width=1\linewidth]{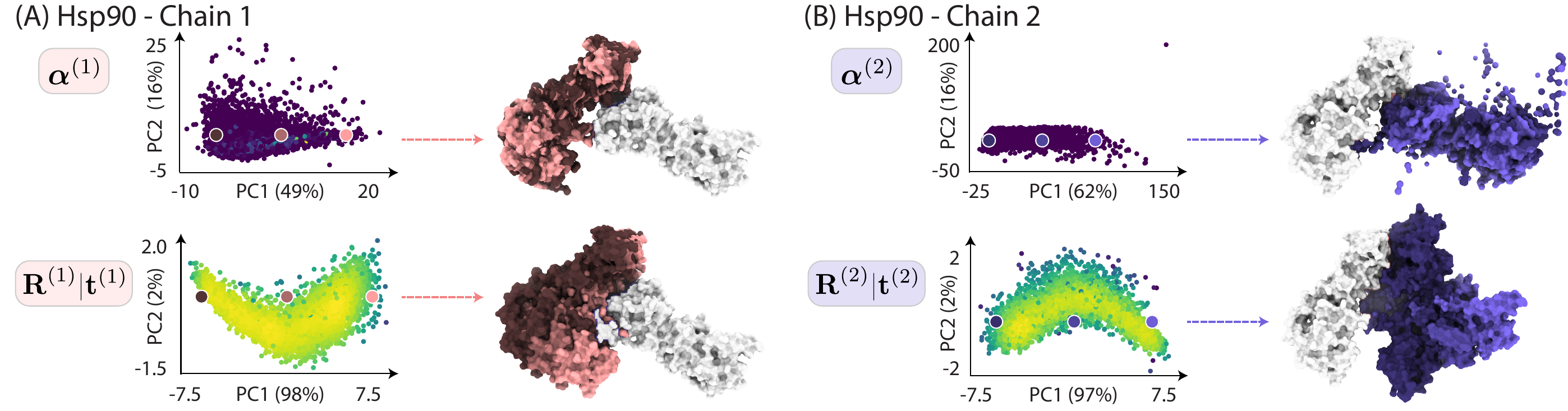}
  \caption{Interpretability of CryoChains's biophysically-grounded latent space for chains 1 (A) and 2 (B) of the heat shock protein Hsp90. Left columns in (A) and (B) show the latent space of cryo-EM images projected on the first two principal components (PCs) of PCA, for normal mode weights ($\boldsymbol{\alpha}
  $) and rigid transformation ($\textbf{R}|\textbf{t}$) per chain. Color represents density (yellow: high density) and explained variance per PC in \%. Right columns show the reconstruction of 3 conformations in 3 shades of red in (A) and blue in (B), sampled along the first PC shown in the corresponding PCA plots. CryoChains's latent space provides an interpretable representation of conformation heterogeneity, e.g., between open $\leftrightarrow$ closed conformations through the rigid transformation.
  }
  \label{fig:cryochains_latent_space}
\end{figure*}

\section{Experiments and Results}
\label{sec:experiments_results}

We show that CryoChains improves 3D reconstruction accuracy while providing an interpretable parameterization of the conformation changes in its latent space.

\subsection{Reconstruction Accuracy}
\label{subsec:quantitative_experiments}




\paragraph{Experiment}
We consider a protein with two conformations
$\textbf{X}^{(0, \text{GT})}$ and $\textbf{X}^{(0)}$, loaded from RCSB PDB 
~\cite{berman_protein_2000}. 
$\textbf{X}^{(0, \text{GT})} \in \mathbb{R}^{3N}$ is a ground truth (GT) reference conformation with total number of atoms $N$. $\textbf{X}^{(0)}  \in \mathbb{R}^{3N}$ is a source reference conformation
used by CryoChains to be fit to 
the images generated with $\textbf{X}^{(0, \text{GT})}$.
We generate a realistic distribution of heterogeneous conformations of this protein using the reference conformations $\textbf{X}^{(c, 0, \text{GT})}$ of its chains,
Eq.~\eqref{eq:conformation_deform} and the following approach simulating normal mode weights per chain. Each
%
weight is computed as
$\boldsymbol{\alpha}^{(c)}_{k} = \sqrt{\frac{N}{K}}d$ 
where $d$ is sampled from the Gaussian mixture model: $d \sim \sum_{m=0}^{1}\phi_{m}\mathcal{N}(\mu_{m}, \sigma^{2}_{m})$
such that $\sum_{m}\phi_{m} = 1$ and
$m \sim \text{Bernoulli}(0.5)$.
After each $\textbf{X}^{(c, \text{GT})}$ is obtained via Eq.~\eqref{eq:conformation_deform}, each chain $c$ is rotated randomly along each axis
in 3D around the chain's CoM.
The final images are rendered using the image formation model of the decoder. 

In practice, we use
the human GABA\textsubscript{B} receptor membrane protein 
and the Hsp90 (heat shock protein of 90 kDa) PDB files. Both proteins have two chains. Fig.~\ref{fig:gaba-synthetic-experiments} shows examples of images generated for each protein.
The PDB files of each protein contains two conformations, closed and open, \texttt{6UO8}/\texttt{6UOA} for GABA\textsubscript{B} and \texttt{2CG9}/\texttt{2CGE} for Hsp90, which play the roles of $\textbf{X}^{(0, \text{GT})}$ and $\textbf{X}^{(0)}$ respectively. 
Concretely, $\textbf{X}^{(0, \text{GT})}=$\texttt{6UO8}/\texttt{2CG9} 
is used to generate synthetic images using 15 normal modes
where $d$ is sampled from a GMM with means $\{0, 2.5\}$ and std $0.25$.
The rotation angles are $\ang{5}$ per axis for \texttt{6UO8} and $\ang{2.5}$ for \texttt{2CG9}.
During training, we deform $\textbf{X}^{(0)}=$\texttt{6UOA}/\texttt{2CGE} 
to fit \texttt{6UO8}/\texttt{2CG9} 
in the images, i.e., fit the open conformation to the closed one.
This setup simulates a real case scenario where we obtain new cryo-EM images, without the GT conformation, and want to reconstruct the 3D structure using an existing PDB file.
Finally, three different levels of noise (SNRs) are added to the images: \{$32$~dB (no noise), $0$~dB, $-20$ dB (close to experimental data)\} following \cite{levyCryoAIAmortizedInference2022}. 
For each noise level, $50,000$ images are generated for the training set, and $5,000$ images for validation and test sets.

\paragraph{Results} We compare the reconstruction accuracy of CryoChains with \cite{nashed_heterogeneous_2022}, the fast, yet biophysically-grounded DL-powered unsupervised heterogeneous reconstruction method in cryo-EM. We also compare CryoChains with its ablation models. We report the root-mean-square deviation (RMSD) between
the predicted and the GT atom coordinates
on GABA\textsubscript{B} and Hsp90 
datasets for each model in Fig.~\ref{fig:gaba-synthetic-experiments}. We highlight that \texttt{N100} refers to \cite{nashed_heterogeneous_2022} using 100 normal modes computed on the whole protein. For both proteins, CryoChains achieves the lowest RMSD, which demonstrates better reconstruction capability. Our ablation study reveals that CryoChains outperforms \texttt{N100}, even when it only relies on per-chain rigid body transformation (\texttt{cRT}). This suggests that introducing rigid transformation of chains is crucial to unlock higher reconstruction accuracy in cryo-EM\textemdash which is precisely the purpose of this paper.
We also show the resulting high-quality 3D reconstructions of our model in Appendix~\ref{sec:additional_experiments}.
We note that the RMSDs of \texttt{N100} and CryoChains
on GABA\textsubscript{B} in Fig.~\ref{fig:gaba-synthetic-experiments}(A) are comparable because GABA\textsubscript{B}'s conformational changes
existing in the dataset do not involve large rigid body transformation
of the chains and thus can be explained well by whole protein NMA only. By contrast, CryoChains significantly improves the RMSDs on Hsp90, because the conformational changes of this protein involve large scale rigid body transformations that can only be captured by our proposed method. CryoChains will indeed provide researchers with maximum benefits in this specific experimental context.


\subsection{Conformational Heterogeneity}
\label{subsec:qualitative_experiments}

\paragraph{Experiment} We perform principal component analysis (PCA) on the latent space of CryoChains trained on images of Hsp90. We demonstrate that CryoChains yields interpretable representations of the protein's 
conformational heterogeneity. We generate
50 Hsp90 conformations by morphing
atom coordinates between Hsp90's two conformations (open and closed) using ChimeraX~\cite{pettersenUCSFChimeraXStructure2021}.
Then, $50,000$ training and $5,000$ test images are generated
with the noise $-20$~dB. We train CryoChains using the open conformation as its source reference $\textbf{X}^{(0)}$. We extract the latent variables corresponding to the test images.
PCA is performed on the latent space of NMA weights ($\boldsymbol{\alpha}$) 
and rigid transformation ($\textbf{R}|\textbf{t}$) per chain.

\paragraph{Results}
In Fig.~\ref{fig:cryochains_latent_space}, the left columns of (A) and (B) show 
the first two principal components of the latent space for NMA weights (top) and rigid transformation (bottom).
The right column depicts 3 conformations, reconstructed from 3 latent points sampled along the first principal component (PC1).
As expected, we observe that the reconstructions from the rigid transformations' principal component vary between the open and closed conformation. 
This demonstrates CryoChains's ability 
to encode the protein's conformation changes
into an interpretable latent space.

\section{Discussions and Conclusion}
\label{sec:discussions_and_conclusions}

We introduced CryoChains to address outstanding challenges in cryo-EM reconstruction.
Our experiments on synthetic datasets show 
that CryoChains is a deep-learning powered approach that improves the 3D reconstruction accuracy of biomolecular structures, while being biophysically interpretable. We hope to open new avenues of research for high-resolution reconstruction of proteins in solution.


\bibliography{cryochains}
\bibliographystyle{icml2023}

\newpage
\appendix
\onecolumn
\section{Normal Mode Analysis}
\label{sec:nma}
Normal mode analysis (NMA) assumes that a set of $N$ atoms are an oscillating system and
represents them with an elastic network model (ENM). 
Noting $\textbf{X} = \left\{\mathbf{r}_j\right\}_{j=1,\ldots, N}$ 
the atomic Cartesian coordinates of the chain, the potential energy $E$ of the ENM is
approximated with a second-order Taylor approximation around a reference conformation 
$\textbf{X}^{(0)}$:
\begin{equation}
    E(\textbf{X}) = \frac{1}{2}\left(\textbf{X}-\textbf{X}^{(0)}\right)^{\top} \textbf{H}\left(\textbf{X}-\textbf{X}^{(0)}\right),
\end{equation}
with $\textbf{H}_{j k} \propto\left(\mathbf{r}_j^{(0)}-\mathbf{r}_k^{(0)}\right)\left(\mathbf{r}_j^{(0)}-\mathbf{r}_k^{(0)}\right)^{\top}$, where $\textbf{H}$ is the Hessian matrix of $E$ in $\textbf{X}^{(0)}$.
Then, eigendecomposition of $\textbf{H}$ is performed
to obtain the eigenvectors $\textbf{U} \in \mathbb{R}^{3N \times 3N}$
which are also referred to as the normal modes.

\section{Image Formation Model}
\label{sec:image_formation_model}
The process of image formation in cryo-EM involves several physical phenomena, 
including pairwise interactions between atoms, interactions between the electron beam and
the molecule's electrostatic potential, or microscope effects.
In CryoChains, the image formation model (Fig.~\ref{fig:cryochains_overview}) is placed in the decoder,
taking in the deformed and rigidly transformed atom coordinates computed from
the dynamics model.
The image formation model is usually written in its ``weak-phase approximation"
where each image $\textbf{I}_i$ in a dataset of $n$ images of $n$ biomolecules is sampled 
according to:
\begin{equation}\label{eq:image_formation}
    \textbf{I}_i = \text{PSF}_i * (\textbf{t}_i + \Pi_{\text{2D}}  \textbf{R}_i) (\textbf{V}_{i})  + \boldsymbol{\epsilon}_i, \text{with }  i=\{1, \ldots n\}.
\end{equation}
\noindent Here, $\textbf{R}_i \in \mathcal{SO}(3)$ is a 3D rotation matrix representing the 3D orientation 
of the volume $\textbf{V}_{i}$ w.r.t. the direction of the electron beam. 
Note that $\textbf{R}_i$ represents a ``global'' 
orientation of the whole biomolecule, as opposed to $\textbf{R}^{(c)}$
inferred in our model.
$\textbf{R}_i$ is obtained using other methods and
assumed that it is given.
This applies to the global translation $\textbf{t}_i$ explained below as well.
The oriented volume is subsequently ``pierced through'' by the electron beam and 
projected onto the detector, which is an operation represented in Eq.~\eqref{eq:image_formation}
by the 2D projection operator $\Pi_{\text{2D}}$. The variable $\textbf{t}_i \in \mathbb{R}^{2}$ represents the 2D translation 
of the projected volume w.r.t. the center of the image. 
The effect of the microscope's lens is modeled through the convolution $*$ of the 2D projection 
with an image-dependent operator $\text{PSF}_i$ called the point spread function of the microscope 
whose parameters can depend on the image. Finally, additional noise $\boldsymbol{\epsilon}_i$ 
is introduced in the generated image, and typically assumed to be Gaussian with zero mean and 
variance $\boldsymbol{\sigma}_i^2$. 

\section{Autoencoder Architecture}
\label{sec:ae}
The encoder $\mathcal{E}_{\theta}$ comprises 4 \texttt{DoubleConvBlock}s
where each of them has 2 blocks of \{\texttt{convolution} - \texttt{max\_pooling} - \texttt{ReLU}\} layers.
The 4 \texttt{DoubleConvBlock}s have \{32, 128, 256, 512\} filters each
and kernels of size $3 \times 3$.
$\mathcal{E}$ embeds the input images into the latent variables,
which branch out to a group of three multilayer perceptrons (MLPs) per chain,
inferring (\textit{i}) per-chain NMA weights $\boldsymbol{\alpha}_{k}$; (\textit{ii}) two 3D vectors for the per-chain rotation; and 
(\textit{iii}) a 3D vector for the per-chain translation.
Each of the MLPs consists of two fully connected layers of sizes \{512, 256\} 
with a \texttt{ReLU} as an activation function in the first layer.
For the second layer, the per-chain rotation MLPs have a $\tanh$ 
and the per-chain translation MLPs do not have an activation function. 


\section{Additional Experiments}
\label{sec:additional_experiments}

\subsection{Qualitative Evaluation}
\begin{figure}[t!]
  \centering
  \includegraphics[width=1\linewidth]{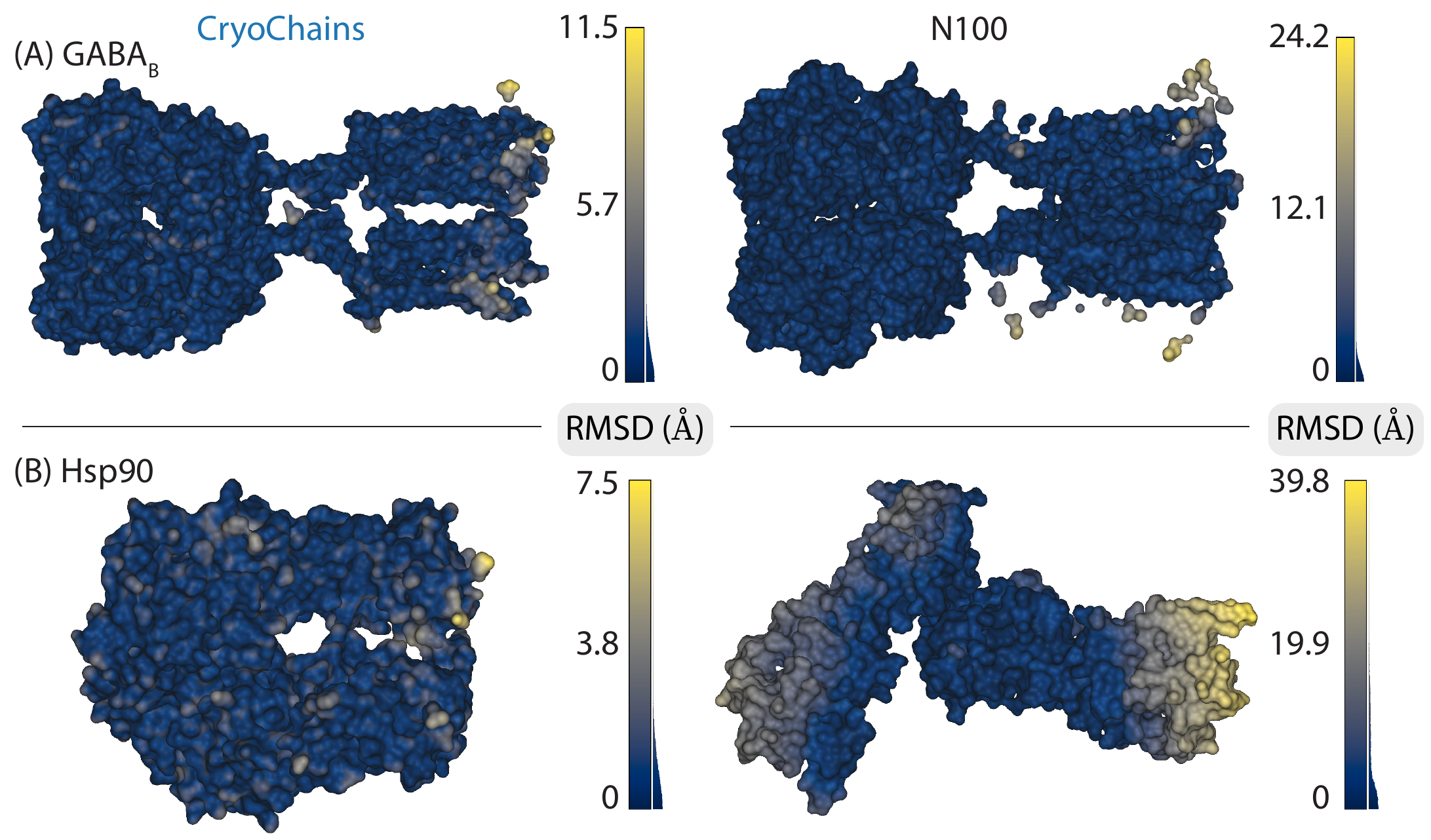}
  \caption{Reconstruction RMSD maps for (A) GABA\textsubscript{B} and (B) Hsp90, using CryoChains (ours) and \texttt{N100}~\cite{nashed_heterogeneous_2022} (whole protein NMA with 100 normal modes). RMSD is computed between the mean conformation of all GT conformations in the dataset and that of all reconstructed conformations. The reconstructed conformation is shown. Next to the colorbar is the histogram for the number of mesh vertices against the errors. The significant improvement in reconstruction for Hsp90 is clearly shown, i.e., the reconstruction using CryoChains correctly matches the closed conformation while that of \texttt{N100} remains open.}
  \label{fig:errormaps}
\end{figure}

Let us visually assess the reconstruction quality of CryoChains
in Fig.~\ref{fig:errormaps}.
We compute the mean conformation of the GT's atom coordinates 
as well as that of the predicted atom coordinates, over the test set.
Then, the Euclidean distances between the vertices of the mean GT and predicted surface meshes
are computed.
We opted to do the above because the reconstruction from our model is in fact the atom coordinates, not voxels or surfaces.
We can observe that the reconstruction of CryoChains is visually better than
that of \texttt{N100}, and the RMSDs are mainly very small except for some extremities of the proteins for both cases
(see also the histograms next to the colorbars).
We can clearly see the merit of CryoChains in the reconstruction of Hsp90, compared to the existing approach.

\end{document}